\documentclass[runningheads]{llncs}
\usepackage[T1]{fontenc}
\usepackage{graphicx}
\usepackage{amsmath,amssymb}
\usepackage{booktabs}
\usepackage{xcolor}
\usepackage{enumitem}
\usepackage{tikz}
\usepackage{hyperref}
\usepackage{listings}
\usepackage{longtable}
\usepackage{microtype}
\usetikzlibrary{arrows.meta,positioning,fit,calc}

\definecolor{softgray}{RGB}{245,245,245}
\definecolor{badred}{RGB}{198,40,40}
\lstdefinestyle{jsonstyle}{basicstyle=\ttfamily\scriptsize,breaklines=true,frame=single,backgroundcolor=\color{gray!5},columns=fullflexible,keepspaces=true}

\newcommand{\data}{\mathcal{D}}
\newcommand{\slice}{\mathcal{S}}
\newcommand{\odd}{\mathcal{O}}

\title{Do Not Forget the Obvious -\\RISC: A Risk-Informed Slice-Coverage Protocol for Safe Autonomous Driving}
\titlerunning{RISC: Risk-Informed Slice Coverage}
\author{Fabian H\"uger}
\authorrunning{Fabian H\"uger}
\institute{CARIAD SE, 38440 Wolfsburg, Germany\\
\email{fabian.hueger@cariad.technology}}

\begin{document}
\maketitle

\begin{abstract}
Aggregate metrics may not fully reflect performance in insufficiently examined high-risk driving conditions. We propose \emph{RISC} (Risk-Informed Slice Coverage), a practical protocol for \emph{risk-guided stress testing} and \emph{coverage-qualified evaluation}. Here, risk-guided stress testing means that a finite audit budget is not sampled uniformly, but directed toward risk-relevant sub-datasets, called \emph{risk slices}; coverage-qualified evaluation means that results are reported together with explicit statements about which slices are sufficiently or insufficiently covered. The protocol translates safety concerns into machine-readable risk slices, uses lightweight signals to tag candidate data, selects a compact audit set by risk, and qualifies the reported results by coverage evidence in each critical slice. An LLM can optionally support this process by surfacing relevant but potentially overlooked conditions for consideration during test planning, thereby helping engineers to \emph{not forget the obvious}. The protocol is model-agnostic and can be applied to perception modules, driving models, or other autonomous-driving subsystems whose safety claims depend on adequate exposure to high-risk conditions. We instantiate the protocol for monocular pedestrian perception in urban driving using 1{,}000 frames from the Zenseact Open Dataset (ZOD), image statistics, and a YOLO-based detector proxy. In this proof-of-concept study, risk-guided selection substantially increases critical failure discovery from 34.0\% under random sampling to 98.5\%. We position the method as a lightweight, assurance-oriented evaluation layer that complements scenario categorization, scenario coverage assessment, and broader testing-and-verification workflows for safe autonomous driving.
\end{abstract}

\keywords{autonomous driving \and stress testing \and safety evaluation \and coverage qualification \and pedestrian detection \and data selection}

\section{Motivation}
Safety-critical autonomous driving systems are evaluated using finite datasets that may not fully represent all relevant operating conditions within their defined operational design domain (ODD), i.e., the operating conditions under which the system is intended to function~\cite{sae2018j3016}. Under these constraints, aggregate metrics may provide an incomplete picture of performance: a subsystem may perform well on average, while conditions associated with potential data-coverage gaps can remain challenging to evaluate. For urban driving, such conditions may include night scenes, partial occlusion by vehicles, small or distant pedestrians, dense crowds, and image degradations such as glare or motion blur. During development and assurance, these conditions should therefore be considered explicitly when evaluating system performance and the available supporting evidence. Their relevance to safety should be assessed in the context of the overall system architecture and its associated safety mechanisms.

Recent progress in large language models creates a practical opportunity for making such evaluation spaces more systematic. LLMs cannot replace safety-critical engineering judgment, but they can help elicit ``obvious'' risk conditions that are easy to overlook in manual test planning, structure them into reviewable slice candidates, and translate them into machine-readable evaluation specifications~\cite{hueger2025agentic}. In this paper, we use this capability as an assurance-support mechanism: the LLM acts as a front-end for surfacing candidate stressors, while all slices, weights, proxy signals, and reported claims remain subject to engineering review and coverage qualification.

We therefore argue for a simple principle:
\begin{quote}
\emph{Do not forget the obvious, and do not interpret results without stating how well these risks are covered.}
\end{quote}

\paragraph{Contributions.}
This paper contributes \emph{RISC} (Risk-Informed Slice Coverage), a practical protocol for risk-guided stress testing and coverage-qualified evaluation of autonomous-driving subsystems. The protocol comprises: (1) a prompt-to-slice step that derives machine-readable stress-test slices from driving safety concerns and applies them to an existing evaluation pool; (2) a risk-weighted selection rule for constructing compact safety-relevant audit subsets; (3) a coverage-qualified reporting step that keeps reported results but explicitly states where evidence is insufficient; and (4) an empirical pedestrian-perception instantiation showing that the protocol increases critical failure discovery from 34.0\% under random sampling to 98.5\% under Risk Top-$K$, thereby exposing safety-relevant hazards for safe autonomous driving.

\section{State of the Art}
Safety assurance in autonomous driving already recognizes that aggregate metrics alone are insufficient. Scenario-based safety-assurance frameworks and recent reviews of scenario metrics emphasize that safety evidence should be structured around scenario coverage, criticality, exposure, completeness, and related forms of contextual evidence rather than being reduced to a single average number. Complementing this perspective, recent testing-and-verification reviews for connected and autonomous vehicles underline that scalable evaluation requires systematic scenario generation, validation approaches that scale to finite engineering budgets, and explicit reasoning about which situations are actually covered by the available evidence.\cite{degelder2025,westhofen2022,alemayehu2025,iso34504}

A second relevant line of work studies evaluation beyond raw perception accuracy. Planner-centric or downstream-sensitive evaluation asks whether perception errors actually matter for subsequent driving behavior. This perspective is closely aligned with safe driving: a missed pedestrian in or near the drivable area, especially under occlusion or at long distance, can require different downstream behavior than an error in an otherwise empty, low-risk scene. Complementary work on uncertainty, calibration, and out-of-distribution behavior in autonomous-driving perception further stresses that reliable deployment depends on knowing where confidence is unwarranted and where data coverage is weak.\cite{philion2020,araujo2024,sicking2022tailored,pintz2022survey}

A third line of work is particularly close to the present paper's proof-of-concept instantiation. Prior work on \emph{safety assurance of machine-learning perception functions} argues that convincing assurance cases require both quantitative performance evidence and structured treatment of ML-specific insufficiencies. Likewise, recent work on \emph{performance limiting factors} for pedestrian detection studies which concrete factors---including distance, occlusion, crowdedness, and visibility effects---systematically contribute to detector misbehavior. These findings motivate our choice to operationalize risk-critical conditions as explicit slices, i.e., named subsets of the data pool, and to bias audit budgets toward the subsets most likely to surface safety-relevant misses.\cite{burton2022,bayzidi2022}

Closely related are automated error-slice discovery approaches such as HiBug2~\cite{chen2025hibug2}, which use task-specific visual attributes and efficient slice enumeration to identify coherent failure subsets for model debugging and repair. Our focus is complementary: rather than discovering new slices or repairing the model, we study how pre-specified, safety-motivated slice proxies can be used to prioritize samples and report audit evidence under a finite evaluation budget.

The gap addressed in this paper is therefore not a new detector or driving model, but an evaluation protocol that contributes to answering three practical questions: how to translate safety concerns into operational slices, how to allocate a finite audit budget toward these slices, and how to report results without overclaiming when the evidence base is thin. We position the proposed protocol as a model-agnostic layer for safety-oriented evaluation, applicable to perception modules today and, in principle, to larger driving-model pipelines as well.

\section{Protocol}
\begin{figure}[t]
\centering
\resizebox{\linewidth}{!}{%
\begin{tikzpicture}[
    font=\scriptsize,
    >=Latex,
    node distance=4mm and 4mm,
    box/.style={
        draw,
        rounded corners=1.8mm,
        fill=softgray,
        align=center,
        text width=27mm,
        minimum height=9mm,
        inner sep=2pt
    },
    arr/.style={->, thick},
    fb/.style={->, thick, dashed}
]

\node[box] (a) {1. Safety concerns\\and latent hazards};
\node[box, right=of a] (b) {2. Risk slices\\and priorities};
\node[box, right=of b] (c) {3. Machine-readable\\rules/spec};
\node[box, right=of c] (d) {4. Candidate tagging\\with lightweight signals};

\node[box, below=10mm of c] (e) {5. Risk scoring\\and Top-$K$ selection};
\node[box, below=10mm of d] (f) {6. Coverage report\\and qualified claims};

\draw[arr] (a) -- (b);
\draw[arr] (b) -- (c);
\draw[arr] (c) -- (d);
\draw[arr] (d.south) -- (e.north);
\draw[arr] (e) -- (f);

\draw[fb] (f.south) -- ++(0,-5mm) -| 
    node[pos=.72, below, font=\scriptsize] {coverage gaps} 
    (b.south);

\end{tikzpicture}%
}
\caption{Schematic overview of the proposed testing workflow. (1) Safety concerns and latent hazards are collected, (2) translated into prioritized risk slices, (3) specified as machine-readable rules, (4) used for candidate tagging with lightweight signals, (5) ranked by risk scoring and Top-$K$ selection, and (6) reported together with coverage qualifications and explicit coverage gaps.}
\label{fig:protocol}
\end{figure}
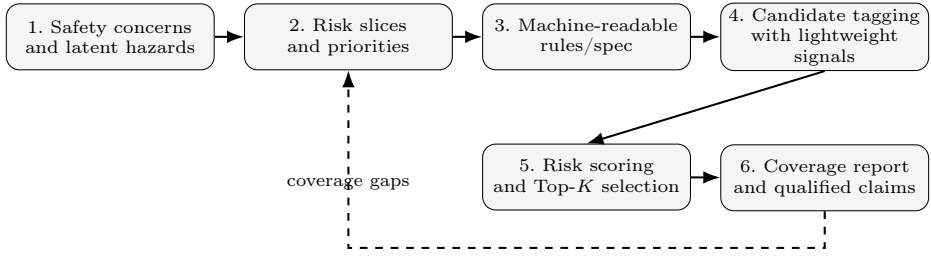

Figure~\ref{fig:protocol} gives a schematic overview of the proposed testing workflow. In short, the protocol translates safety concerns into interpretable risk slices, operationalizes them through lightweight tagging rules, ranks candidate samples by risk, and reports the resulting evaluation together with explicit coverage qualifications.

Let $\data=\{x_1,\ldots,x_N\}$ be a pool of frames, clips, or episodes from an operational design domain $\odd$. A risk slice $S_i \subseteq \data$ is an interpretable subset of this pool, for example night scenes, small pedestrians, or pedestrians close to vehicles. We define $K$ such slices, $\slice=\{S_1,\ldots,S_K\}$, with binary indicators $z_i(x)\in\{0,1\}$ for $i=1,\ldots,K$. Each slice receives a weight
\begin{equation}
    w_i \propto \mathrm{Severity}_i \times \mathrm{Exposure}_i \times \mathrm{Detectability}_i^{-1} \times \mathrm{Overreliance}_i.
\end{equation}
Here, severity describes the potential consequence of a missed hazard, exposure describes how plausible the condition is in the intended ODD, detectability describes how difficult the condition is expected to be for the evaluated function, and overreliance captures the risk that good average performance leads to excessive trust in the function. Samples are ranked by
\begin{equation}
    r(x)=\sum_{i=1}^{K} w_i z_i(x)+\lambda u_{\mathrm{proxy}}(x),
\end{equation}
where a higher $r(x)$ means that sample $x$ has higher priority for inclusion in the audit set. The optional term $u_{\mathrm{proxy}}(x)$ can encode lightweight evidence of uncertainty, image degradation, detector confidence, or disagreement between multiple candidate taggers or model variants. The top-$B$ samples form a stress-test or audit set.

\subsection{Risk-slice elicitation from safety concerns}
The protocol begins with a structured safety prompt that asks which hazards should matter for evaluation. The purpose is not to generate a universal ontology, but to create an inspectable initial taxonomy of risk-critical slices that can later be reviewed and operationalized by engineers. A standard large language model is useful in this step because it already encodes many common, obvious failure modes from public autonomous-driving and perception knowledge; these suggestions are treated as candidates, not as authoritative safety requirements. For example, a prompt may ask for pedestrian-perception hazards in urban driving and yield slices such as \emph{night/low light}, \emph{vehicle occlusion}, and \emph{small or distant pedestrians}, together with severity rationale, plausible exposure, expected detectability, overreliance risk, and candidate tagging heuristics.

\subsection{Machine-readable specification}
The elicited slices are translated into a compact, structured specification, for example in JSON- or YAML-like form. This machine-readable layer is essential because it decouples the safety reasoning from the implementation and enables reusability across curation, sampling, monitoring, and audit pipelines.

\subsection{Candidate tagging with lightweight signals}
Each slice is approximated with lightweight signals, for example timestamps, image statistics, detector outputs, or simple geometry heuristics. For instance, low-light candidates can be tagged by low mean brightness and a high dark-pixel ratio, while a vehicle-occlusion proxy can be tagged by small distances or overlap between detected person and vehicle boxes. The goal is not perfect semantic labeling, but a low-cost candidate-generation stage that can be audited and improved later.

\subsection{Risk-weighted selection}
Once candidate tags are available, each sample receives a risk score. The score combines the selected slice weights with optional proxy terms such as image degradation, uncertainty, or disagreement between taggers or model variants. The original evaluation pool is not used only as-is because finite audit budgets can be dominated by easy or frequent cases; Top-$K$ selection instead concentrates manual inspection and detailed analysis on samples expected to be rich in safety-critical failures. Random and heuristic baselines are retained as controls.

\subsection{Coverage diagnostics and qualified reporting}
The final step is not to suppress results, but to qualify them. If a slice remains undercovered, the evaluation still reports the measured performance, but it additionally states that the evidence base for that slice is weak. This distinction is important: the protocol is intended to prevent overinterpretation, not to block all reporting.

\section{Pedestrian-Perception Instantiation for Safe Driving}
We instantiate the protocol for monocular pedestrian perception in urban driving. Image statistics are used to tag low light, blur, and glare, while predictions from a frozen detector proxy are used to derive candidate tags for small/distant pedestrians, crowds, and vehicle-occlusion proxies. These detector-derived tags are used only for candidate generation and risk-based sampling; ZOD ~\cite{alibeigi2023zod} annotations are used for the subsequent performance evaluation. This separation is important because proxy tags are not ground truth and may share failure modes with the evaluated detector. In this framing, pedestrian-perception failures are interpreted as failures to surface hazards that should trigger more defensive downstream driving behavior.

\subsection{Why pedestrian perception matters for safe driving}
Reliable pedestrian detection is relevant to the development of safe automated driving systems. The perception task can be particularly challenging in situations involving partial occlusion, small or distant pedestrians, or dense interaction zones. Examples include a pedestrian emerging from behind a vehicle, a distant pedestrian at a crossing, or multiple pedestrians in a crowded scene. These conditions should therefore be considered explicitly when evaluating pedestrian-perception functions and the available supporting evidence.

\subsection{Prompting procedure and proof-of-concept scope}
The slice taxonomy in this paper was produced as a proof-of-concept prompting exercise rather than a universal standard. In a first stage, a structured safety prompt was given to GPT-5.5 Thinking to elicit candidate slices, rationales, and operational heuristics. Details of the prompt and prompting setup can be found in Appendix A. In a second stage, the resulting list was manually reviewed and reduced to a compact subset that could be implemented with the available image statistics, detector outputs, and audit budget. The purpose was to demonstrate a practical workflow from safety concern to operational slice, not to claim that the resulting prompt or slice set is universally optimal.

\begin{table}[t]
\centering
\caption{Compact slice specification and implementation signals used in the proof-of-concept study.}
\label{tab:slices}
\begin{tabular}{lll}
\toprule
ID & Slice & Operational signal used in prototype \\
\midrule
S1 & Night / low light & mean brightness and dark-pixel ratio \\
S2 & Vehicle occlusion proxy & person--vehicle proximity or box overlap\\
S4 & Small/distant pedestrians & person-box height or image-area ratio \\
S5 & Motion blur & Laplacian-variance sharpness below threshold \\
S6 & Glare/backlight & bright-pixel ratio and high contrast \\
S7 & Dense crowd & number of detected persons \\
\bottomrule
\end{tabular}
\end{table}
In Table~\ref{tab:slices}, S3 is intentionally absent because the crosswalk-interaction slice suggested during prompting could not be implemented reliably with the available lightweight signals in this proof-of-concept. The detector-derived slices should therefore be read as proxy candidates for stress testing, not as complete semantic labels.

\subsection{Reasoning behind the weights}
Table~\ref{tab:weights} reports the additive slice weights used in the prototype. The weights in this paper are not learned parameters and are not normalized probabilities; they are scoring weights for ranking samples. They are a compact operationalization of a safety argument. We assign higher weight to slices that combine: high potential severity if missed, plausible exposure in urban traffic, low detectability by the perception stack, and elevated risk of human or system overreliance.

Concretely, the highest weight is assigned to \textbf{S2 vehicle occlusion proxy} because hidden pedestrians can emerge suddenly from behind parked or moving vehicles and are both frequent and safety-critical in urban scenes. \textbf{S1 night / low light} receives a high weight because low visibility degrades detection while also encouraging misplaced confidence if aggregate daytime performance is overgeneralized. \textbf{S4 small/distant pedestrians} remains highly relevant because safe action often depends on early recognition of distant, low-pixel-footprint pedestrians. \textbf{S6 glare/backlight} and \textbf{S5 blur} are weighted lower in this proof-of-concept not because they are unimportant, but because they appeared less frequently in the available 1{,}000-frame subset. \textbf{S7 crowd} captures interaction density and ambiguity, which can increase both miss probability and downstream risk.

\begin{table}[t]
\centering
\caption{Risk weights used for additive risk scoring in the detector-assisted proof-of-concept.}
\label{tab:weights}
\begin{tabular}{llc}
\toprule
Slice & Meaning & Weight \\
\midrule
S1 & Night / low light & 0.19 \\
S2 & Vehicle occlusion proxy & 0.23 \\
S4 & Small/distant pedestrians & 0.15 \\
S5 & Motion blur & 0.05 \\
S6 & Glare/backlight & 0.11 \\
S7 & Dense pedestrian crowd & 0.08 \\
\bottomrule
\end{tabular}
\end{table}

\section{Experiment: Risk-Guided Selection}
\subsection{Setup}
We process 1{,}000 monocular front-camera frames and compare three sampling strategies with budget $K=200$: uniform random sampling, a simple human heuristic baseline, and risk-guided Top-$K$ selection. The human heuristic baseline is a simple non-learned rule-based selection intended to represent manually chosen obvious stressors, rather than a trained risk model. The detector configuration uses YOLOv8n, the nano variant of YOLOv8, with detector input size 640 and confidence threshold 0.25; the original images are resized for detector inference and are not assumed to be square. Blur is measured by the variance of the image Laplacian, a common sharpness proxy: lower values indicate stronger blur. We flag the lowest 10\% of frames by this sharpness score, which yields an effective threshold of 14.73 in this dataset. Detection performance is evaluated against pedestrian annotations from the Zenseact Open Dataset (ZOD). Predicted and annotated pedestrian boxes are matched using an IoU threshold of $\geq 0.5$; objects marked as unclear are excluded.

\begin{figure}[t]
\centering
\IfFileExists{fig_risk_score_distribution_xlabels_same_size_cropped.pdf}{\includegraphics[width=.72\linewidth]{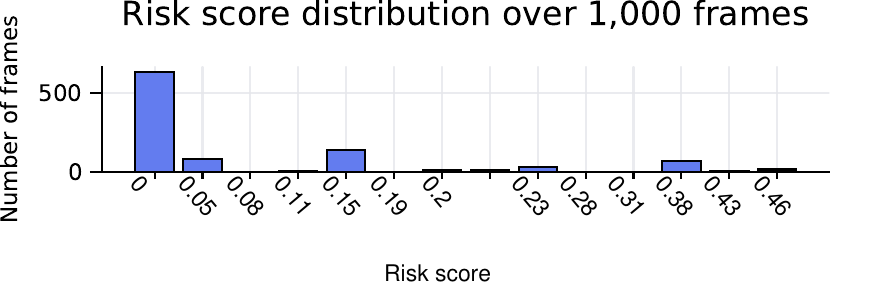}}{\fbox{Risk-score histogram placeholder.}}
\caption{Risk score distribution over 1{,}000 processed frames. Most frames are not part of any detected risk slice, while high-risk multi-slice frames form a small tail.}
\label{fig:riskdist}
\end{figure}

\subsection{Risk and coverage distribution}
\begin{figure}[t]
\centering
\IfFileExists{fig_slice_coverage_methods.pdf}{\includegraphics[width=.9\linewidth]{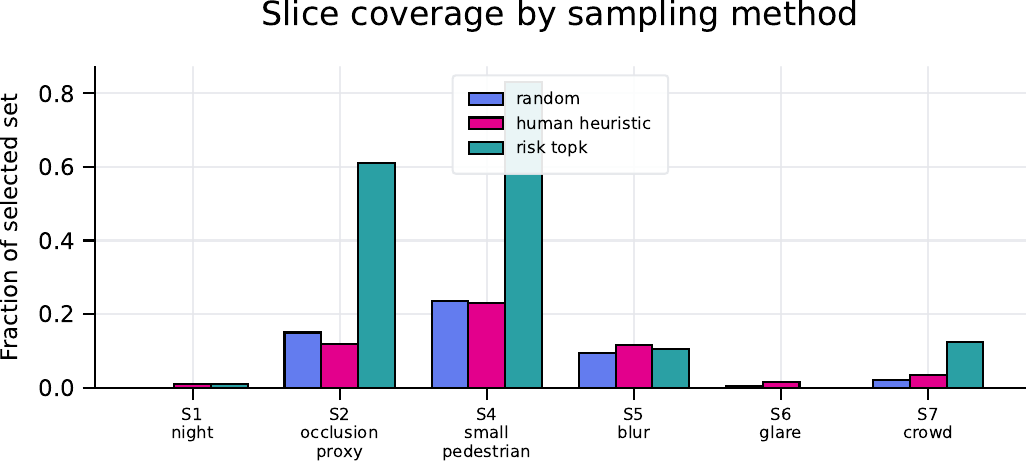}}{\fbox{Slice-coverage plot placeholder.}}
\caption{Slice coverage by sampling method. Risk-guided selection concentrates the audit budget on occlusion, small-pedestrian, and crowd slices.}
\label{fig:coverage-methods}
\end{figure}

As shown in Figure~\ref{fig:riskdist}, the frame pool is strongly imbalanced: 634 of 1{,}000 frames receive risk score 0, while only 20 frames reach scores of 0.43 or higher. As can be seen in Table~\ref{tab:coverage-results} and Figure~\ref{fig:coverage-methods}, Risk Top-$K$ enriches detector-derived stress slices: S2 occlusion proxy rises from 15.0\% under random sampling to 61.0\%, S4 small pedestrians from 23.5\% to 83.0\%, and S7 crowd from 2.0\% to 12.5\%. Night and glare do not increase substantially because these conditions are rare in the available 1{,}000-frame pool and can only be enriched if enough candidate frames are present.

\begin{table}[t]
\centering
\caption{Slice coverage comparison. The Manifest column reports prevalence in the full 1{,}000-frame candidate pool; Random, Human heuristic, and Risk Top-$K$ report prevalence within the selected $K=200$ audit sets.}
\label{tab:coverage-results}
\begin{tabular}{lrrrr}
\toprule
Slice & Manifest & Random & Human heuristic & Risk Top-$K$ \\
\midrule
S1 night & 0.2\% & 0.0\% & 1.0\% & 1.0\% \\
S2 occlusion proxy & 12.2\% & 15.0\% & 12.0\% & \textbf{61.0\%} \\
S4 small pedestrian & 24.9\% & 23.5\% & 23.0\% & \textbf{83.0\%} \\
S5 blur & 10.0\% & 9.5\% & 11.5\% & 10.5\% \\
S6 glare & 0.3\% & 0.5\% & 1.5\% & 0.0\% \\
S7 crowd & 2.6\% & 2.0\% & 3.5\% & \textbf{12.5\%} \\
\bottomrule
\end{tabular}
\end{table}

\subsection{Performance metrics}
Coverage alone does not measure perception quality. We therefore evaluate the same frozen detector on all three selected test sets using pedestrian annotations from ZOD and report the standard set-level counts and rates---precision, recall, and miss rate---together with two audit-oriented metrics that are directly tied to the proposed protocol.

\paragraph{Standard metrics.}
Precision, recall, and miss rate are computed in the standard way from $TP$, $FP$, and $FN$ after IoU-based matching. Miss rate is emphasized because the practical concern is how many pedestrians remain undetected in the selected audit sets.

\paragraph{Critical Failure Discovery Rate (CFDR@$K$).}
Let $\mathcal{A}_K$ denote the selected audit set and let $c(x)=1$ if frame $x$ contains at least one false negative and belongs to at least one non-zero-risk slice, and $c(x)=0$ otherwise. We compute
\[
\mathrm{CFDR}@K = \frac{1}{K}\sum_{x\in\mathcal{A}_K} c(x).
\]
CFDR@$K$ therefore measures the fraction of selected audit frames that actually surface missed pedestrians in risk-relevant conditions. This metric is intentionally conditioned on the risk-slice concept and should not be interpreted as an unbiased general-performance metric. Instead, it answers the budget-oriented audit question: \emph{how often does the selected audit set surface failures in risk-relevant conditions?}

\paragraph{Risk-Weighted Failure Discovery (RWFD@$K$).}
CFDR@$K$ treats all critical-failure frames equally. RWFD@$K$ further weights each critical-failure frame by its risk score $r(x)$ and normalizes by the audit budget:
\[
\mathrm{RWFD}@K = \frac{1}{K}\sum_{x\in\mathcal{A}_K} r(x)c(x).
\]
In this experiment, risk scores range from 0 to 0.46; examples include 0.15 for a small-pedestrian-only frame, 0.23 for an occlusion-proxy-only frame, and 0.46 for frames combining small pedestrians, crowds, and occlusion proxies. RWFD@$K$ therefore captures not only whether failures are found, but whether they are found in higher-priority regions of the risk space.

Table~\ref{tab:performance-metrics} reports the resulting set-level performance and audit-yield metrics.
\begin{table}[t]
\centering
\caption{Detector performance and failure-discovery metrics on the three selected test sets. GT denotes pedestrian instances in ZOD; matching uses IoU $\geq 0.5$.}
\label{tab:performance-metrics}
\scriptsize
\begin{tabular}{lrrrrrrrrr}
\toprule
Test set & GT & TP & FP & FN & Prec. & Rec. & Miss & CFDR@$K$ & RWFD@$K$ \\
\midrule
Random ($K=200$) & 2029 & 92 & 69 & 1937 & 57.1\% & 4.5\% & 95.5\% & 34.0\% & 7.56\% \\
Human heuristic ($K=200$) & 1892 & 100 & 58 & 1792 & 63.3\% & 5.3\% & 94.7\% & 35.0\% & 7.21\% \\
Risk Top-$K$ ($K=200$) & 2560 & 326 & 207 & 2234 & 61.2\% & 12.7\% & 87.3\% & 98.5\% & 27.81\% \\
\bottomrule
\end{tabular}
\end{table}

\paragraph{Interpreting performance across sets.}
The evaluation confirms that the risk-guided set is substantially harder and more failure-dense than the baselines. Compared with random sampling, Risk Top-$K$ increases recall from 4.5\% to 12.7\% because the selected set contains more pedestrian instances and more frames in which the detector produced at least one pedestrian prediction; this should not be interpreted as improved detector robustness. The miss rate nevertheless remains high at 87.3\%, indicating that the selected frames expose many missed pedestrians. Most importantly for stress testing, CFDR@$K$ rises from 34.0\% for random sampling and 35.0\% for the heuristic baseline to 98.5\% for Risk Top-$K$; the risk-weighted failure discovery rate increases from 7.56\% to 27.81\%. Thus, the main result is not that the detector becomes better on the risk-selected set. Rather, risk-guided selection makes the fixed audit budget more informative by selecting frames in which missed pedestrians occur more often in safety-relevant conditions.

Table~\ref{tab:risk-topk-slice-performance} further qualifies the Risk Top-$K$ result by breaking down detector performance inside the selected risk slices. The occlusion-proxy and small-pedestrian slices contain many pedestrian instances and retain high miss rates, indicating that the selected audit set indeed concentrates on difficult safety-relevant conditions. The crowd slice shows higher recall but still exposes substantial remaining misses. These slice-level results should not be interpreted as independent robustness estimates, because frames may belong to multiple slices and rare slices such as night or glare remain undercovered. Instead, the table identifies which covered risk slices drive the audit yield and where additional evidence would be needed.

\begin{table}[t]
\centering
\caption{Slice-level detector performance within the Risk Top-$K$ set. We report slices with sufficient selected frames and available ground-truth pedestrians for meaningful interpretation; S1 night and S6 glare are omitted here because they are too rare or absent in the selected subset. The hardest selected slices contain many pedestrian instances and high miss rates, which is precisely the intended stress-test behavior.}
\label{tab:risk-topk-slice-performance}
\scriptsize
\begin{tabular}{lrrrrrrrr}
\toprule
Slice & Frames & GT & TP & FP & FN & Prec. & Rec. & Miss \\
\midrule
S2\_occlusion\_proxy & 122 & 1660 & 207 & 146 & 1453 & 58.6\% & 12.5\% & 87.5\% \\
S4\_small\_pedestrian & 166 & 2210 & 298 & 178 & 1912 & 62.6\% & 13.5\% & 86.5\% \\
S5\_blur & 21 & 185 & 23 & 17 & 162 & 57.5\% & 12.4\% & 87.6\% \\
S7\_crowd & 25 & 574 & 147 & 68 & 427 & 68.4\% & 25.6\% & 74.4\% \\
\bottomrule
\end{tabular}
\end{table}

\section{Coverage-Qualified Reporting}
For each slice $S_i$, we report its prevalence in the candidate pool and in the selected audit sets. The reporting rule is:
\begin{quote}
\textbf{Coverage qualification.} Report the measured result, but explicitly state when a risk-critical slice remains undercovered and therefore provides only limited evidence.
\end{quote}
The experiment illustrates why this qualification is necessary. Risk-guided Top-$K$ improves S2, S4, and S7 coverage substantially, but it does not automatically cover all relevant slices. In particular, glare (S6) and night (S1) remain rare in the 1{,}000-frame subset. Therefore, the reported results are informative, but they should not be read as comprehensive evidence for driving safety outside the covered conditions.

\section{Discussion and Limitations}
RISC is an assurance-support protocol, not a replacement for scenario engineering, verification, or safety-case development. Its main contribution is to separate three concerns that are often mixed in aggregate evaluation: \emph{what to test} through risk slices, \emph{what to prioritize} through risk-weighted selection, and \emph{what can be claimed} through coverage-qualified reporting.

The pedestrian-perception experiment is only an exemplary instantiation of the protocol, not a comprehensive benchmark of detector robustness. It illustrates how safety concerns can be translated into operational slice proxies, how a finite audit budget can be directed toward higher-risk samples, and how the resulting findings can be reported without overclaiming.

The proof-of-concept remains limited by approximate detector-derived slice tags, partial coupling between proxy tagging and detector evaluation, and an intentionally unbalanced Risk Top-$K$ audit set. Consequently, the reported performance numbers should be interpreted as audit-yield diagnostics rather than unbiased robustness estimates. Rare slices such as night and glare also remain undercovered in the available 1{,}000-frame pool.

These limitations reinforce the purpose of coverage-qualified reporting: RISC does not turn limited evidence into broad safety claims, but makes the boundary of the available evidence explicit. Future work should combine global Risk Top-$K$ selection with slice-balanced quotas, random controls, temporal driving scenes, and downstream planning metrics.

\section{Conclusion}
We presented \emph{RISC}, a compact risk-informed slice-coverage protocol for risk-guided stress testing and coverage-qualified evaluation in safe autonomous driving. In an empirical pedestrian-perception experiment, risk-guided selection enriched occlusion, small-pedestrian, and crowd-related stress slices compared with random and heuristic baselines, while increasing critical failure discovery from 34.0\% to 98.5\%. The resulting evaluation combines coverage, perception performance, failure discovery, and explicit coverage qualifications, enabling more transparent reporting of evaluation evidence and its limitations for autonomous driving systems.

\section*{Acknowledgement}
The research leading to these results is funded by the German Federal Ministry for Economic Affairs and Energy within the project ``Safe AI Engineering -- Sicherheitsargumentation bef\"ahigendes AI Engineering \"uber den gesamten Lebenszyklus einer KI-Funktion''. The author would like to thank the consortium for the successful cooperation.

\newpage
\appendix
\section*{Appendix}
\section{Prompting Setup and Scope}
\label{app:prompting-setup}
The prompt used in this paper is a \emph{proof-of-concept prompt}, not a generally valid template. Its purpose was to test whether a modern foundation model can help bootstrap an inspectable slice taxonomy for autonomous-driving safety evaluation.

The prompting workflow had two stages. In Stage~1, \textbf{GPT-5.5 Thinking} was asked to generate risk-relevant slices, rationales, priorities, and candidate heuristics for pedestrian perception in urban driving. The emphasis was on operationalizability rather than completeness: each slice had to be approximable from timestamps, image statistics, detector outputs, or simple heuristics. In Stage~2, the resulting slice set was manually reviewed, pruned, and aligned with the signals available in the actual prototype pipeline. The final subset retained in the paper should therefore be read as a compact demonstration set for this study, not as a domain-independent ontology.

\section{Stage-1 Prompt for Risk-Slice Elicitation}
\label{app:stage1-prompt}
The following prompt was used to elicit the initial risk-critical slice taxonomy. It is included to make the slice construction process inspectable and reproducible.

\begin{quote}\small
You are a Safety and ML Assurance Engineer for a camera-based pedestrian detection system deployed in urban driving environments.

\textbf{Goal.}
Identify, structure, and prioritize risk-relevant data slices, i.e., operational design domain subspaces, for pedestrian detection, such that they can later be used for: (a) dataset curation and focused sampling, (b) safety monitoring and triggering, (c) audit and assurance documentation, and (d) automated tagging and processing in downstream pipelines.

\textbf{Context and assumptions.}
\begin{itemize}[leftmargin=*]
 \item Task: pedestrian detection, either bounding boxes or presence detection, using monocular RGB camera input.
 \item Operational domain: urban Germany, EU context, mixed-density city traffic, speeds between 0--60 km/h, frequent pedestrian interactions.
 \item Environmental conditions: day/night cycles, weather variations, and typical European urban infrastructure such as crosswalks, sidewalks, and intersections.
 \item System context: human-in-the-loop operation possible; risk of operator overreliance on model outputs should be explicitly considered.
 \item Constraints: outputs must be structured to enable simple tagging heuristics without requiring large-scale labeling or complex models.
 \item Important: all slices must be defined in a way that allows them to be approximated using simple signals such as metadata, timestamps, image statistics, detector outputs, or lightweight heuristics.
\end{itemize}

\textbf{Task.}
\begin{enumerate}[leftmargin=*]
 \item Generate a list of 15--25 risk-relevant data slices for pedestrian detection.
 \item For each slice, provide a compact operational specification with: slice ID and name; short definition; risk rationale; risk factors including severity, exposure, detectability, and overreliance trigger; tagging and measurement heuristics; typical failure modes; suggested evaluation criterion; priority; and assumptions.
 \item Additionally provide: (A) a Top-8 shortlist of critical slices, (B) a compact subset suitable for a proof-of-concept study, (C) ten high-risk combinatorial slices, and (D) a simple weighting scheme for slice importance using
 \[
 w_i \propto \mathrm{Severity}_i \times \mathrm{Exposure}_i \times \frac{1}{\mathrm{Detectability}_i} \times \mathrm{Overreliance}_i.
 \]
 \item After the human-readable section, generate a machine-readable representation of all slices in valid JSON with fields for ID, name, definition, numeric risk factors, priority, heuristics, failure modes, acceptance criterion, tags, and combinable slices. Optionally also provide YAML suitable for pipeline configuration.
\end{enumerate}

\textbf{Constraints and style.}
Avoid vague categories; every slice must be operationalizable. Do not assume proprietary signals; use only camera input and common metadata. Focus on real-world urban driving conditions in Germany. Emphasize safety-critical pedestrian scenarios. Use precise engineering language and ensure consistency between human-readable and machine-readable outputs.
\end{quote}

\end{document}